\documentclass[letterpaper]{article} 

\usepackage{xcolor}

\usepackage{aaai25}

\usepackage{times}  
\usepackage{helvet}  
\usepackage{courier}  
\usepackage[hyphens]{url}  
\usepackage{graphicx} 
\usepackage{natbib}  
\usepackage{caption} 
\usepackage{algorithm}
\usepackage{algorithmic}

\usepackage{booktabs}

\usepackage{newfloat}
\usepackage{listings}
\DeclareCaptionStyle{ruled}{labelfont=normalfont,labelsep=colon,strut=off} 
\floatstyle{ruled}
\newfloat{listing}{tb}{lst}{}
\floatname{listing}{Listing}
\title{\raisebox{-1.5ex}{\includegraphics[height=1.8em]{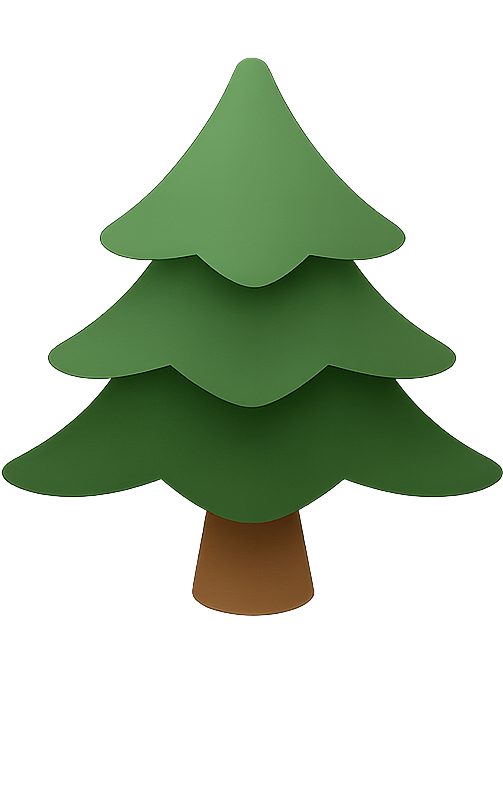}}\hspace{0.2em}\normalfont\bfseries AgriGPT: a Large Language Model Ecosystem for Agriculture}

\author{
Bo Yang$^{1}$, Yu Zhang$^{1}$, Lanfei Feng$^{2}$, Yunkui Chen$^{2}$, Jianyu Zhang$^{1}$, Xiao Xu$^{1}$,\\
Nueraili Aierken$^{1}$, Yurui Li$^{1}$, Yuxuan Chen$^{1}$, Guijun Yang$^{3}$, Yong He$^{4}$,\\
Runhe Huang$^{5}$, Shijian Li$^{1}$\thanks{Corresponding author} \\
$^{1}$College of Computer Science and Technology, Zhejiang University, China \\
$^{2}$College of Software Technology, Zhejiang University, China \\
$^{3}$School of Geological Engineering and Geomatics, Chang'an University, China \\
$^{4}$College of Biosystems Engineering and Food Science, Zhejiang University, China \\
$^{5}$Faculty of Computer and Information Sciences, Hosei University, Japan \\
\texttt{\{boyang30, 22421173, jianyu.zhang, 3200105334, nureli, liyr, yuxuan\_chen, yhe, shijianli\}@zju.edu.cn}, \\
\texttt{\{22451116, 22351048\}@zju.edu.cn}, \texttt{yanggj@chd.edu.cn}, \texttt{rhuang@hosei.ac.jp}
}
\usepackage{bibentry}

\begin{document}

\maketitle

\begin{abstract}
Despite the rapid progress of Large Language Models (LLMs), their application in agriculture remains limited due to the lack of domain-specific models, curated datasets, and robust evaluation frameworks. To address these challenges, we propose \textbf{AgriGPT}, a domain-specialized LLM ecosystem for agricultural usage.
At its core, we design a multi-agent scalable data engine that systematically compiles credible data sources into \textbf{Agri-342K}, a high-quality, standardized question-answer (QA) dataset. Trained on this dataset, AgriGPT supports a broad range of agricultural stakeholders, from practitioners to policy-makers.
To enhance factual grounding, we employ \textbf{Tri-RAG}, a three-channel Retrieval-Augmented Generation framework combining dense retrieval, sparse retrieval,and multi-hop knowledge graph reasoning,  thereby improving the LLM reasoning reliability.
For comprehensive evaluation, we introduce \textbf{AgriBench-13K}, a benchmark suite comprising 13 tasks with varying types and complexities. Experiments demonstrate that AgriGPT significantly outperforms general-purpose LLMs on both domain adaptation and reasoning.
Beyond the model itself, AgriGPT represents a modular and extensible LLM ecosystem for agriculture, comprising structured data construction, retrieval-enhanced generation, and domain-specific evaluation. This work provides a generalizable spectrum for developing scientific and industry specialized LLMs. All models, datasets, and code will be released to empower agricultural communities, especially in underserved regions, and promote open, impactful research.
\end{abstract}

\begin{figure*}[!htb]
  \centering
  \includegraphics[width=\textwidth,trim=0 4cm 0 0, clip]{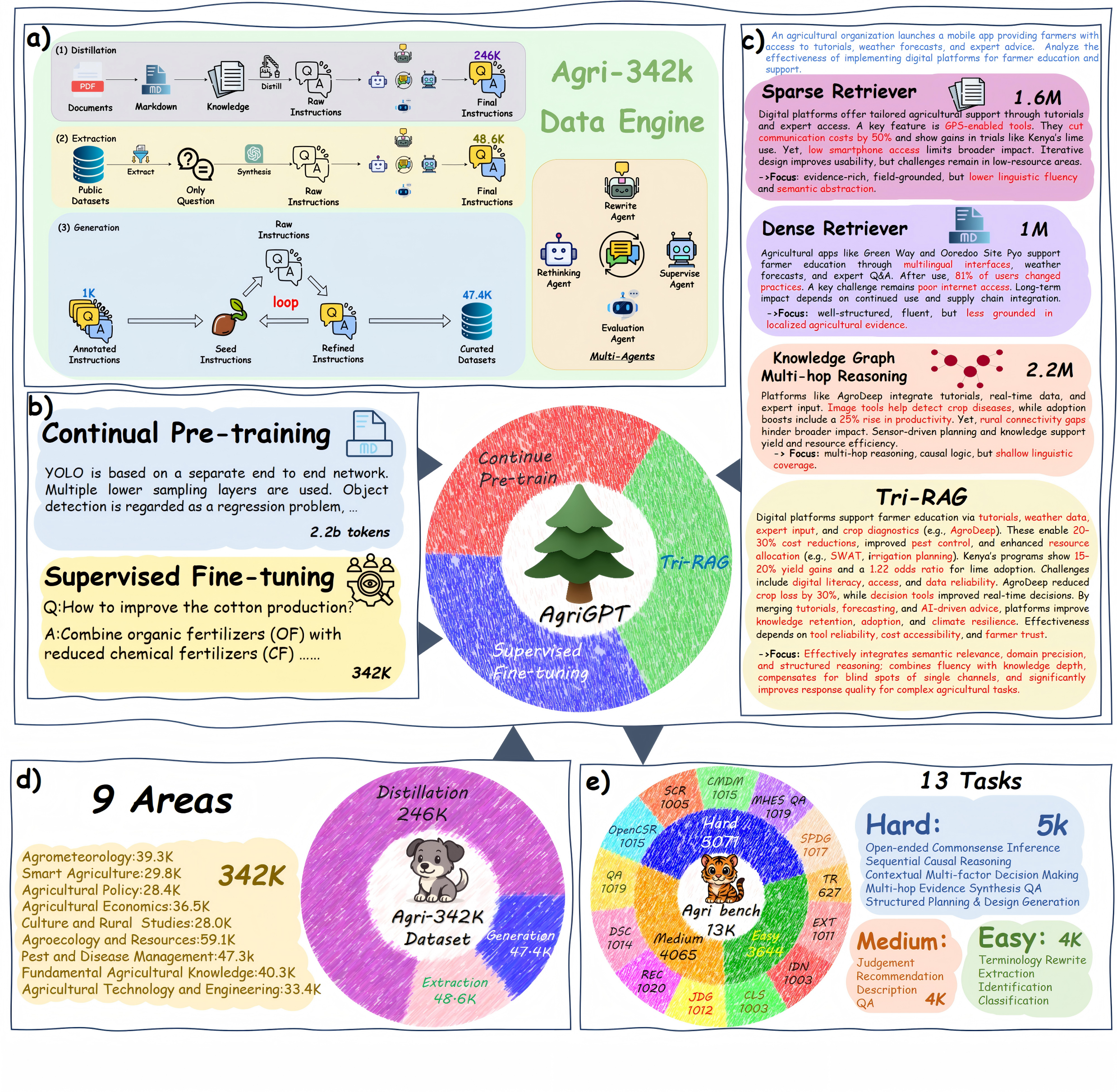}
  \caption{The AgriGPT Ecosystem:  a). AgriGPT Data Engine: the 3 pipelines to construct Agri-342K dataset b). illustrating the model training workflow (continual pretraining and supervised fine-tuning) c). Tri-RAG inference and ablation: highlighting multi-path gains over single-path baselines d). the Agri-342K dataset with a broad topic spectrum e). the AgriBench-13K benchmark design}
  \label{fig:overview}
\end{figure*}

\section{Introduction}

Agriculture is central to global food security and sustainability \cite{short2023scalable,di2016enhancing,swaminathan2001food,rozenstein2024data,kamilaris2018deep,wolfert2017big,liakos2018machine,foley2011solutions,clapp2020food}, yet remains underrepresented in AI research. Effective decisions rely on integrating diverse knowledge from crop science to market signals. Addressing this gap is critical, especially for smallholder farmers facing knowledge constraints \cite{godfray2010food,altieri2009agroecology,rockstrom2017sustainable,calicioglu2019future,fan2020role,klerkx2020dealing,vanlauwe2014sustainable,rivera2009extension}.

Recent progress in LLMs \cite{chen2024survey,jiang2023structgpt,zha2023tablegpt,fang2023knowledge,zhao2023survey,achiam2023gpt} shows promise in enabling multi-modal reasoning, but their direct application in agriculture is limited by sparse terminology coverage, lack of retrieval grounding, and domain-specific reasoning gaps \cite{rezayi2022agribert,yang2024shizishangpt}. Agricultural data are fragmented, unstructured, and domain experts are scarce \cite{wigboldus2016systemic}. Moreover, successful deployment requires not only accuracy, but local relevance, interpretability, and infrastructure-awareness \cite{ghanem2015agriculture}.

Several domain-specific LLMs exist-e.g.,AgroGPT  \cite{awais2025agrogpt}, 
AgroLLM   \cite{samuel2025agrollm},
AgriLLM   \cite{didwania2024agrillm},
and AgriBERT \cite{rezayi2022agribert}, as well as task-specific applications such as LLM-based QA systems \cite{he2024enhancing} and plant disease detection models \cite{zhao2024implementation}.In other domains, notable examples include BioGPT \cite{luo2022biogpt}, 
MedPaLM \cite{tu2024towards}, 
LegalGPT \cite{shi2024legalgpt}. However, most of these models lack generative reasoning capabilities, depend on complex pipelines, or are not open-source.
Tools like ShizishanGPT \cite{yang2024shizishangpt} rely on complex retrieval pipelines. In contrast, AgriGPT is a unified, open ecosystem integrating scalable instruction curation and reasoning.

Unlike prior works such as MAG-v \cite{sengupta2024mag}, BioInstruct \cite{tran2024bioinstruct}, and Self-Instruct \cite{wang2022self}, we construct \textbf{Agri-342K}, a large instruction dataset using our multi-agent  \textbf{data engine} generation. Our retrieval module  \textbf{(Tri-RAG)} combines dense, sparse, multi-hop knowledge graph reasoning strategies \cite{lewis2020retrieval,xiong2020answering,rackauckas2024rag,sanmartin2024kg,peng2024graph,arslan2024survey,jin2024long} to address factuality and reasoning gaps. We also introduce \textbf{AgriBench-13K}, a benchmark covering factual QA, diagnostic reasoning, , multi-hop reasoning and several tasks.

By releasing all code, data, and benchmarks, we aim to lower barriers to agricultural AI deployment. Grounded in global development and AI-for-social-good discourse \cite{vinuesa2020role,crawford2021atlas,shi2020artificial}, AgriGPT supports equitable access to intelligent tools for farmers, and lays a foundation for scalable, socially impactful LLM applications in agriculture.

To sum up, our contributions are as follows:

\begin{itemize}

\item \textbf{Agri-342K dataset:} we develop a multi-agent data engine that enables scalable, high-quality curation of agricultural knowledge to the creation of 342K instruction data, along with a multilingual version to support cross-lingual capabilities.

\item \textbf{AgriBench-13K Benchmark Suite:} A multi-task, multi-level benchmark for agricultural LLMs, including Mini-AgriBench600 and its multilingual variant for lightweight and cross-lingual evaluation.

\item \textbf{AgriGPT Ecosystem:} We propose the first comprehensive LLM ecosystem for agriculture and open-source its model, dataset, and benchmark, while providing a scalable and transferable framework—via our multi-agent engine and Tri-RAG—for building domain-specific LLMs in other real-world fields.

\end{itemize}

\section{AgriGPT}

In this section, we present the AgriGPT ecosystem (Figure~\ref{fig:overview}). Our methods are derived as follows: we first introduce our data engine, a multi-agent data creation pipeline that constructs the Agri-342K instruction dataset and AgriBench-13K. Then, we describe the continual pretraining stage for domain adaptation. This is followed by supervised fine-tuning to align the model with task-specific objectives. Next, we present Tri-RAG, a retrieval-augmented inference module that integrates dense retrieval, sparse retrieval,and multi-hop knowledge graph reasoning to enhance factual accuracy and reasoning. Finally, we detail the development of AgriBench-13K, a multi-task benchmark for evaluating model performance across diverse agricultural tasks and difficulty levels.

\subsection{Data Engine}

We define the scope of agricultural knowledge by organizing it into 9 major thematic domains, each representing a core area of agricultural science and practice.
To ensure comprehensive coverage within each domain, we manually curated a list of over 600 sub-area keywords, which were used to guide document retrieval, filtering, and categorization.
Based on these keywords, we collected a wide range of credible, domain-relevant documents from research papers, technical manuals, and textbooks. Thematic distribution of these documents across the 9 domains is shown in Table~\ref{tab:category_keywords}.

As illustrated in Figure~\ref{fig:overview}(a), we develop a modular and scalable data creation engine that constructs instructions. The data engine consists of 3 pipelines to systematically aggregate variety of data sources into unified data format. For publicly available data, we use distillation to process credible documents as raw instructions. For existing public question datasets \cite{SivaResearch_agri_2024,KisanVaani_agriculture_qa_2024}, we employ general domain LLM (DeepSeek-R1-671B \cite{guo2025deepseek}) to generate raw answers. For scaling up data synthesis, we initiate a loop by 1K human verified instructions as seed to continuously generate new raw instructions, while accepting a portion of curated data to enrich the seed and improve synthetic quality.

\begin{figure*}[!htbp]
  \centering
  \includegraphics[width=\textwidth]{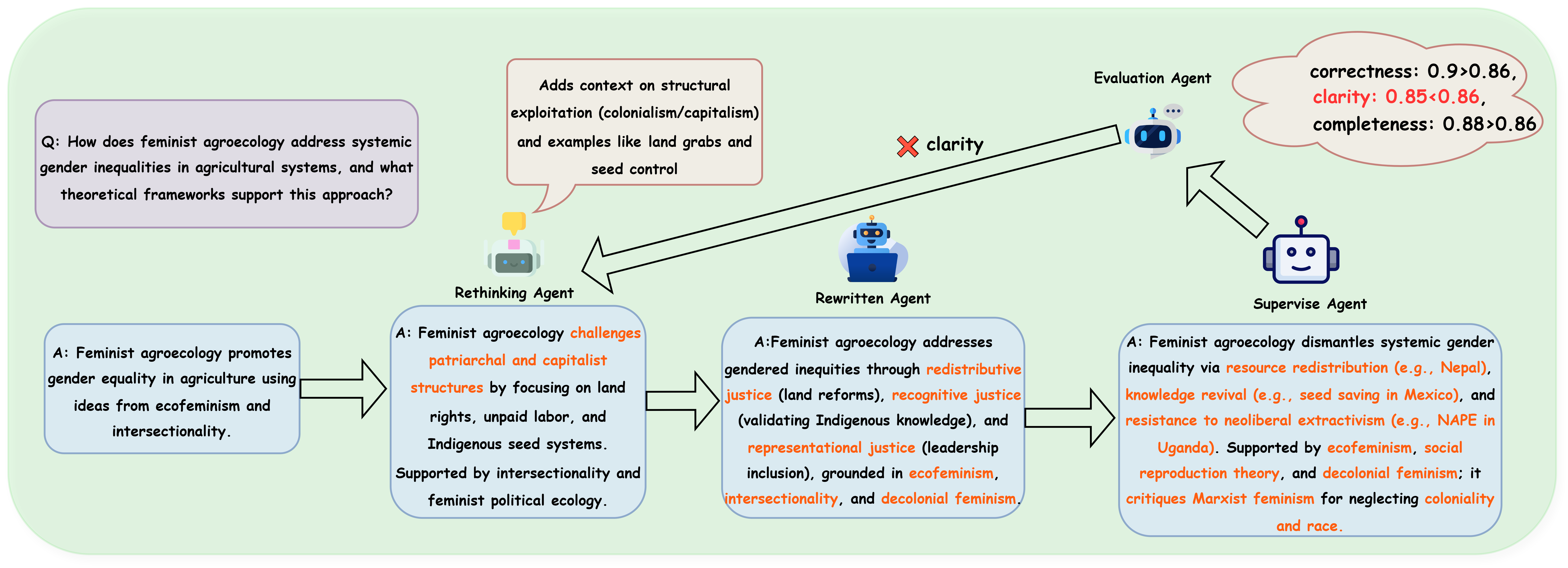}
  \caption{Workflow: Multi-Agent Framework for Ensuring Instruction Data Quality}
  \label{fig:Workflow}
\end{figure*}

For all the raw instructions from 3 pipelines, we design 4 collaborative agents powered by DeepSeek-R1-671B to finalize them into logical, diverse and factual grounded instructions: As illustrated in Figure~\ref{fig:Workflow}. First, the \textbf{Rethinking Agent} revisits each QA pair by simulating alternative reasoning paths and exploring different semantic perspectives. This process helps uncover logical gaps, improve coverage, and enhance the diversity and robustness of the content. Next, the \textbf{Rewrite Agent} ensures stylistic coherence and linguistic normalization by paraphrasing the text, standardizing format and tone, and enforcing the use of domain-specific terminology, all while aligning with established instructional prompting practices. The \textbf{Supervise Agent} then acts as a semantic validator, checking alignment between the question and its source context, verifying factual correctness, and filtering hallucinated or off-topic content. Finally, the \textbf{Evaluation Agent} applies both rule-based and model-driven scoring metrics to assess each QA pair across dimensions such as coherence, informativeness, and accuracy. Based on these evaluations, only the highest-quality samples are retained, ensuring the overall reliability and instructional value of the dataset.

\begin{table}[ht]
\centering

\begin{tabular}{l r}
\toprule
\textbf{Areas} & \textbf{Num} \\
\midrule
Fundamental Agri Knowledge         & 11306  \\
Pest and Disease Management                & 17856  \\
Agroecology and Natural Resources          & 24947  \\
Agri Technology and Engineering    & 21000  \\
Smart Agri, AI \& Computing         & 21062  \\
Agri Economics                     & 17741  \\

Meteorology, Remote Sensing                & 29884  \\
Agricultural Policy and Governance         & 17914  \\
Life, Culture and Rural Studies            & 20304  \\
\midrule
\textbf{Total}                                 & \textbf{182014} \\
\bottomrule
\end{tabular}
\caption{Thematic distribution of agricultural documents}
\label{tab:category_keywords}
\end{table}

The statistics of 3 pipelines are as follows:

\begin{itemize}

\item \textbf{Distillation:} We process approximately 182k research papers and 591 foundational books using MinerU to extract clean markdown-formatted content. All documents are first passed through a rule-based preprocessing pipeline, including structure-aware parsing, noise removal (e.g., boilerplate, references, page numbers), and keyword-based filtering to ensure a high signal-to-noise ratio. To further improve content relevance, we apply BM25 filtering guided by our 600+ sub-area taxonomy. The retained passages are then processed through our \textit{collaborative multi-agent framework}, which orchestrates prompt construction, content paraphrasing, and iterative quality control. This distillation process transforms raw literature into high-quality, domain-specific QA pairs that reflect scientific depth and terminological precision.

\item \textbf{Extraction:} We collect agriculture related questions from public QA datasets and encyclopedic sources. Since many entries lack complete answers or follow inconsistent formats, we reformat them into instruction-answer pairs using prompt-based rewriting. These samples are then passed through the \textit{collaborative multi-agent framework}, where the answers are regenerated to align with the agricultural language conventions and evaluated for coherence, correctness and topical alignment.

\item \textbf{Generation:} We manually compose a set of about 1K expert-written seed prompts covering diverse agricultural topics and task types. These prompts serve as anchors for iterative prompt expansion. Using the \textit{collaborative multi-agent framework}, we generate rephrased, extended, and related question variants along with corresponding answers through multi-round prompting. The outputs are deduplicated, reviewed, and filtered to ensure factual accuracy and diversity. We also pay special attention to expert agreement and noise control during annotation to ensure high-quality supervision, resulting in a robust supplemental QA set.
\end{itemize}

The full pipeline yields over 342K reasoning-augmented QA pairs covering almost all areas of agriculture. Our data engine offers a reusable framework for constructing domain-specific instruction datasets at scale.

\subsection{Continual Pretraining Stage}

As shown in Figure~\ref{fig:overview} b), before fine-tuning on Agri-342K, we perform a continual pretraining stage on Qwen3-8B \cite{yang2025qwen3}. We adopt a LoRA-based \cite{hu2022lora} approach to learn agricultural terminology and linguistic patterns from the corpus while maintaining its general language capabilities.
As an extension of the model’s original pretraining, this process helps AgriGPT absorb specialized vocabulary (e.g., crop species names, disease terminology) and factual knowledge from agricultural literature.
The continual pretraining phase mitigate catastrophic forgetting of general knowledge and effectively adapt to the agricultural domain, laying a solid foundation for subsequent supervised fine-tuning.

\begin{figure*}[!htbp]
  \centering
  \includegraphics[width=\textwidth]{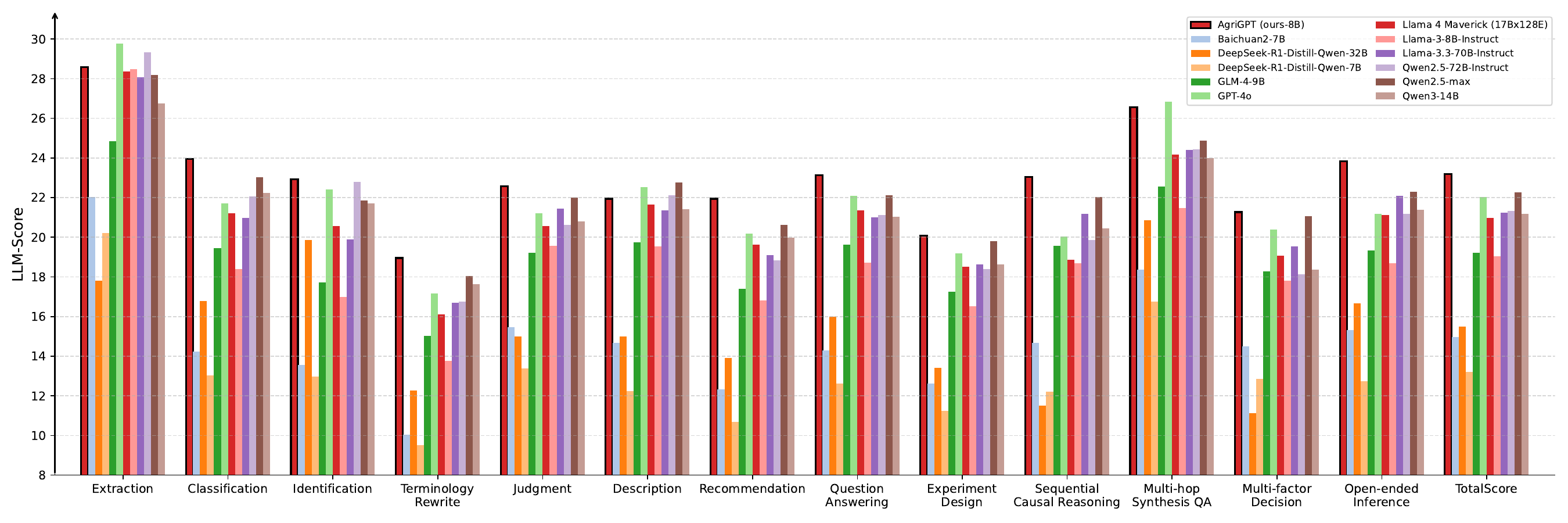}
  \caption{LLM-Based Evaluation of AgriGPT and Other Models across 13 Tasks of AgriBench-13K and Total Score}
  \label{fig:13tasks llm evaluation}
\end{figure*}

\subsection{Supervised Fine-Tuning Stage}

As shown in Figure~\ref{fig:overview} b), after continual pretraining, we perform Supervised Fine-Tuning (SFT) on the Agri-342K dataset created by the data engine.
In this stage, the model learns to follow a question and generate a correct answer in our dataset format.
We fine-tune the model’s parameters (including the LoRA adapter weights and, optionally, the top layers of the base model) using a standard language modeling objective to maximize the probability of the correct answer given the question with cross entropy loss. 
The fine-tuning process exposes the model to a wide variety of agriculture queries: from simple factual questions to complex explanatory prompts—and their corresponding answers. 
As a result, AgriGPT learns to produce high-quality answers that are both informative and domain-appropriate.
The SFT stage solidifies the model’s ability to perform as an agricultural assistant, aligning its generation style with the QA format and content of our curated dataset.

\subsection{Tri-RAG Stage}

To enhance AgriGPT’s ability to handle complex agricultural queries, we propose a three channel Retrieval-Augmented Generation (Tri-RAG) framework that incorporates three complementary information sources at inference time.
The design centers on combining dense semantic matching, domain-focused filtering, and structured reasoning to provide rich and diverse external context for generation.

The first channel builds a dense semantic index over the credible research papers and agricultural textbooks from data sources, allowing the model to retrieve document that are semantically aligned with the query.
The second channel applies a BM25-based sparse retrieval strategy to extract high-relevance agricultural fragments from the same corpus, ensuring that the retrieved content is both targeted and domain-specific.
The third channel further processes the BM25-filtered content by extracting approximately 2 million factual triples in the form of (entity, relation, entity), which are then used to construct a knowledge graph for multi-hop graph-based reasoning.

Each retrieval channel contributes uniquely: the dense retriever offers semantically fluent matches, the sparse retriever ensures domain-specific precision, and the graph-based retriever enables structured, multi-hop reasoning. During inference, outputs from all three channels are merged and re-ranked using a composite relevance scoring function that accounts for both retrieval confidence and content diversity. This process filters out redundant or low-quality information, ensuring that only the most informative and complementary evidence is selected. The top-ranked passages and/or factual triples are then integrated with the original query to construct an augmented prompt, which is fed into AgriGPT. Leveraging its pretraining on agricultural corpora and instruction fine-tuning, the model can effectively utilize this contextual information during generation.

As shown in Figure~\ref{fig:overview} c), We clearly demonstrate the performance differences between using a single-channel RAG and the Tri-RAG. By combining dense semantic alignment, domain-specific retrieval, and graph-based reasoning, the Tri-RAG framework empowers AgriGPT to produce responses that are factually grounded, contextually relevant, and logically coherent. This multi-channel design significantly enhances the model's performance on complex, knowledge-intensive agricultural tasks that require both precise retrieval and multi-step reasoning.

\begin{table}[ht]
\centering
\begin{tabular}{l r}
\toprule
\textbf{Easy Tasks} & \textbf{Num} \\
\midrule
Extraction             & 1,011 \\
Classification         & 1003 \\
Identification         & 1003 \\
Terminology Rewrite    & 627 \\
\midrule
\textbf{Medium Tasks} & \textbf{Num} \\
\midrule
Judgment               & 1,012 \\
Description            & 1,014 \\
Recommendation         & 1,020 \\
Question Answering     & 1,019 \\
\midrule
\textbf{Hard Tasks} & \textbf{Num} \\
\midrule
Experiment Design        & 1,017 \\
Sequential Causal Reasoning               & 1,005 \\
Multi-hop Evidence Synthesis QA           & 1,019 \\
Contextual Multi-factor Decision       & 1,015 \\
Open-ended Commonsense Inference             & 1,015 \\
\bottomrule
\end{tabular}
\vspace{0.5em}
\caption{Task distribution of AgriBench-13K}
\label{tab:agribench_task_dist}
\end{table}

\begin{table*}[t] 
\centering
\resizebox{\textwidth}{!}{ 
\begin{tabular}{lcccccc}
\toprule
\textbf{Model} & \textbf{BLEU} & \textbf{Meteor} & \textbf{Rouge-1} & \textbf{Rouge-2} & \textbf{Rouge-L} & \textbf{LLM-Score} \\
\midrule
Baichuan2-7B&1.57&14.13&20.55&4.41&19.64&14.96\\
DeepSeek-R1-Distill-Qwen-7B&1.91&13.60&20.22&4.59&19.34&13.21\\
GLM-4-9B&9.46&31.23&27.55&6.33&26.31&19.22\\
Meta-Llama-3-8B-Instruct&7.81&28.56&25.43&5.78&24.28&19.05\\
DeepSeek-R1-Distill-Qwen-32B&3.06&15.03&21.52&5.66&20.50&15.49\\
Llama-3.3-70B-Instruct&8.25&30.07&26.25&6.31&25.04&21.23\\
Qwen3-14B&7.63&25.27&\underline{29.20}&\underline{8.20}&\underline{27.90}&21.19\\
Qwen2.5-72B-Instruct&11.29&33.52&28.31&7.23&27.04&21.32\\
GPT-4o&8.41&27.97&27.58&6.97&26.27&22.02\\
Llama 4 Maverick (17Bx128E)&7.37&28.37&26.76&6.60&25.55&20.98\\
Qwen2.5-max&\underline{12.38}&\underline{38.14}&28.70&7.10&27.30&\underline{22.27}\\
AgriGPT (ours)&\textbf{16.52}&\textbf{44.06}&\textbf{31.60}&\textbf{9.63}&\textbf{30.32}&\textbf{23.20}\\
\bottomrule
\end{tabular}
}
\label{tab:performance}
\end{table*}

\begin{table*}[t] %
\centering
\resizebox{\textwidth}{!}{ 
\begin{tabular}{ccccccc}
\toprule
\textbf{Correctness} & \textbf{Match ability} & \textbf{Fluency} & \textbf{Coherence} & \textbf{Relevance} & \textbf{Logical Consistency} & \textbf{Completeness} \\
\midrule
2.26&2.06&2.37&2.02&2.19&2.03&2.02\\
1.99&1.78&2.14&1.86&1.97&1.78&1.69\\
3.01&2.60&2.94&2.61&2.78&2.64&2.64\\
2.93&2.59&2.91&2.62&2.79&2.62&2.59\\
2.37&2.18&2.38&2.12&2.26&2.10&2.08\\
3.27&2.90&3.22&2.88&3.10&2.91&2.95\\
3.24&2.76&3.22&2.93&\underline{3.23}&2.90&2.90\\
3.30&2.94&3.19&2.93&3.11&2.94&2.91\\
3.41&3.05&\underline{3.27}&3.03&3.19&\underline{3.05}&3.02\\
3.26&2.89&3.12&2.87&3.06&2.91&2.89\\
\underline{3.53}&\underline{3.08}&3.27&\underline{3.07}&3.22&3.04&\underline{3.07}\\
\textbf{3.59}&\textbf{3.16}&\textbf{3.44}&\textbf{3.19}&\textbf{3.36}&\textbf{3.21}&\textbf{3.25}\\
\bottomrule
\end{tabular}
}
\caption{Comparison of AgriGPT with general LLMs. \textbf{Bold} and \underline{underlined}  indicate best and second-best performance.}
\label{tab:total performance}
\end{table*}

\subsection{AgriBench-13K}

As shown in Figure~\ref{fig:overview} e), to systematically assess the capabilities of LLMs in agriculture, we present AgriBench-13K, a comprehensive benchmark specifically designed for evaluating LLMs in agricultural contexts. It comprises 13 representative task types,  these tasks reflect a wide range of language understanding and reasoning challenges encountered in real-world agricultural scenarios. The distribution of task types is summarized in Table~\ref{tab:agribench_task_dist}.

The benchmark construction begins with domain experts defining 13 distinct task categories. Based on 9 major agricultural domains and over 600 sub-area labels, a total of 585 seed prompts are created, each corresponding to a core problem instance within a specific task and sub-area combination. These seeds are expanded using our data engine, which iteratively generates large-scale candidate question–answer pairs through cyclic sampling and multi-round refinement. All samples are subsequently de-duplicated and manually reviewed to ensure quality and consistency. To ensure fairness, we strictly separate the benchmark from the training data (Agri-342K) and apply similarity-based filtering to avoid data leakage.

The resulting AgriBench-13K benchmark includes 12,780 high-quality samples, with broad topical and task coverage. For efficient evaluation, we also uniformly sample a lightweight subset, Mini-AgriBench600. As the first large-scale standardized benchmark in the agricultural domain, it offers a unified framework for evaluating the performance of agricultural LLMs and provides a foundation for advancing model development, training strategies, and system-level innovations in agriculture.

\begin{figure}[htbp]
  \centering
  \includegraphics[width=0.9\linewidth, trim=0cm 0cm 0cm 0cm, clip]{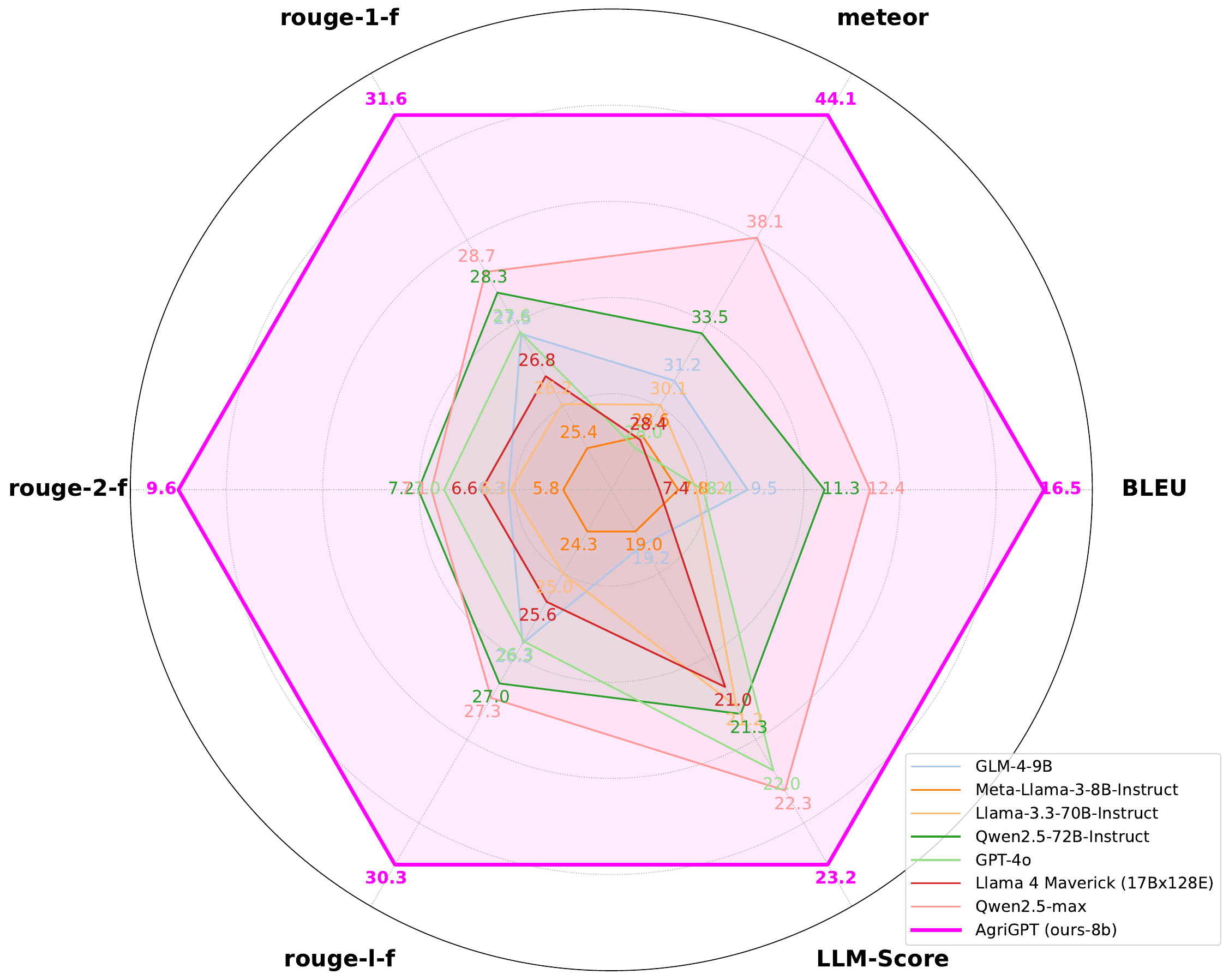}
  \caption{Performance comparison of AgriGPT and other models on AgriBench-13K}
  \label{fig:agrigpt_radar}
\end{figure}

\section{Result}
\subsection{Comparative Experiments} 

As illustrated in Table~\ref{tab:total performance} and Figure~\ref{fig:agrigpt_radar}, We evaluate AgriGPT against eleven representative LLMs: 
Baichuan2-7B \cite{yang2023baichuan}, 
DeepSeek-R1-Distill-Qwen-7B \cite{deepseekai2025deepseekr1incentivizingreasoningcapability},
GLM-4-9B \cite{glm2024chatglm}, 
Meta-Llama-3-8B-Instruct \cite{dubey2024llama3},
DeepSeek-R1-Distill-Qwen-32B \cite{deepseekai2025deepseekr1incentivizingreasoningcapability}, 
Llama-3.3-70B-Instruct \cite{meta2024llama3.3}, 
Qwen3-14B \cite{yang2025qwen3},   
Qwen2.5-72B-Instruct \cite{qwen2.5-72b},    
GPT-4o \cite{achiam2023gpt}, 
Llama 4 Maverick (17B×128E) \cite{meta2024llama4}, 
and Qwen2.5-max \cite{qwen2.5-max}.   
Overall and task-specific results are presented detailed in Appendix.

To the best of our knowledge, there are currently may not have open-source domain-specific LLMs for agriculture such as AgroGPT(Awais et al. 2025), AgroLLM (Samuel et al. 2025), AgriLLM (Didwania et al. 2024), AgriBERT (Rezayi et al.2022). AgriGPT is the first open-source model in this vertical domain, and thus we compare it only with general-purpose LLMs. Overall and task-specific results are presented in detail in the Appendix.

We adopt a dual evaluation strategy combining automatic metrics and LLM-based scoring. Traditional metrics such as BLEU \cite{papineni2002bleu}, METEOR \cite{denkowski2014meteor}, and ROUGE \cite{lin2004rouge} quantify surface-level accuracy, while Qwen2.5-72B serves as an expert evaluator to score outputs across seven qualitative dimensions: Correctness, Match Ability, Fluency, Coherence, Relevance, Logical Consistency, and Completeness. Each dimension is scored from 0 (failure) to 5 (excellent), accompanied by a confidence score (0 to 1). The final LLM-Score is computed as a confidence-weighted average, improving robustness by prioritizing high-certainty judgments. 

This hybrid approach ensures a balanced assessment of both lexical fidelity and deeper semantic quality. It demonstrates that AgriGPT not only delivers accurate responses but also generates coherent, logically consistent, and contextually relevant content—critical for real-world agricultural applications.

From the results, we observe the following. (1) \textbf{AgriGPT consistently achieves top scores} across both automatic and LLM-based evaluation strategies, highlighting its overall effectiveness. (2) \textbf{it also attains best ratings across all LLM-evaluated dimensions}, accompanied by high confidence scores, indicating both strong generation quality and reliable semantic alignment. (3) \textbf{the agreement between automatic metrics and LLM-based evaluations} reflects the balanced design of AgriGPT, which performs well at both surface-level accuracy and deeper reasoning. (4) \textbf{despite its relatively compact size}, AgriGPT outperforms many larger closed-source commercial LLMs, striking a strong balance between performance and efficiency. Its lightweight design enables deployment in low-resource regions and on constrained devices, enhancing real-world and social impact.

Figure~\ref{fig:13tasks llm evaluation}
shows the LLM evaluation results of AgriGPT and other purpose-specific models across all 13 tasks in AgriBench-13K. AgriGPT outperforms across the majority of tasks, with particularly strong results in multi-hop reasoning and sequential understanding, demonstrating its versatility and robustness in complex agricultural scenarios.

\subsection{Generalization Evaluation}

To ensure AgriGPT avoids domain overfitting, we evaluate its generalization by comparing performance against the base model on several public general-domain benchmarks, including MMLU\cite{hendrycks2020measuring,mathew2021docvqa} , ARC \cite{allenai:arc}, and OpenBookQA \cite{OpenBookQA2018}—three widely-used benchmarks that span diverse knowledge areas such as science, humanities, and logical reasoning, and are standard for evaluating language model generalization. The results (Table~\ref{tab:generalization})  show that AgriGPT performs comparably to Qwen3-8B, with only marginal differences within 0.5 points across datasets. This demonstrates that our domain adaptation enhances agricultural capabilities without compromising general reasoning and language skills.

\begin{table}[ht]
\centering
\resizebox{\linewidth}{!}{ %
\begin{tabular}{lccc}
\toprule
\textbf{Model} & \textbf{MMLU} & \textbf{ARC} & \textbf{OpenBookQA} \\
\midrule
Qwen3-8B(base) & 85.87\% & 97.56\% & 91.77\% \\
AgriGPT(ours) & 85.84\% & 97.49\% & 91.20\% \\
\bottomrule
\end{tabular}
}
\caption{Generalization to public benchmarks}
\label{tab:generalization}
\end{table}

\subsection{Multilingual Evaluation} 

To evaluate multilingual capabilities, we translated Agri-342K into multiple languages to construct Agri-342K-Multilingual and conducted instruction tuning on AgriGPT. For efficient testing, we sampled 600 examples from AgriBench-13K to create Mini-AgriBench600, then translated it into multiple languages (Mini-AgriBench600-Multilingual). As shown in Table~\ref{tab:multilingual}, the model achieves reasonable BLEU and Meteor scores on Chinese and Japanese, indicating effective transfer of instruction-following ability across languages.

\begin{table}[ht]
\centering
\small
\setlength{\tabcolsep}{25pt}

\begin{tabular}{lcc}
\toprule
\textbf{Language} & \textbf{BLEU} & \textbf{Meteor} \\
\midrule
English  & 16.52 & 40.16 \\
Chinese  & 17.80 & 48.42 \\
Japanese & 16.69 & 45.30 \\
\bottomrule
\end{tabular}
\caption{Multilingual Evaluation}
\label{tab:multilingual}
\end{table}

\subsection{Ablation Study} 
We conduct an ablation study to assess the effects of domain-specific training and RAG. As shown in Table~\ref{tab:Ablation Study}, both components individually improve BLEU and Meteor scores, while their combination yields the best performance, highlighting their complementary benefits.

\begin{table}[ht]
\centering
\small
\setlength{\tabcolsep}{8pt} 
\begin{tabular}{lcc}
\toprule
\textbf{Model} & \textbf{BLEU} & \textbf{Meteor} \\
\midrule
Qwen3-8B  & 12.53 & 41.51 \\
Qwen3-8B + RAG & 13.39 & 42.06 \\
Qwen3-8B + domain training  & 16.29 & 43.76 \\
Qwen3-8B + domain training + RAG & 16.42 & 44.15 \\
\bottomrule
\end{tabular}
\caption{Ablation Study of Domain training and RAG }
\label{tab:Ablation Study}
\end{table}

\subsection{Potential Social Impact and Limitations }
AgriGPT has the potential to significantly enhance agricultural productivity and decision-making in underserved rural regions. By enabling intelligent question answering, policy support, and real-time analysis in local agricultural contexts, it empowers farmers, extension workers, and policymakers with accessible, domain-specific knowledge. We further demonstrate its deployability by achieving 44.15 token/s inference speed on a single RTX 4090 GPU, enabling cost-effective edge deployment.This supports scalable usage in low-resource settings, helping reduce knowledge inequality, promote sustainable practices, and improve food security outcome.

AgriGPT currently has three key limitations: it only supports text input without multimodal capabilities; its training data lacks diversity due to reliance on formal sources; and it does not explicitly handle regional dialects. Future work will focus on adding image and sensor inputs, incorporating informal texts and farmer dialogues, and expanding dialect coverage to improve real-world applicability.

\section{Conclusion}

We present AgriGPT, a domain-specific LLM designed to support complex agricultural tasks. Starting with a scalable, collaborative multi-agent data engine and a Tri-RAG framework, we obtain a high-quality Agri-342K instruction dataset that captures diverse agricultural knowledge. To systematically assess model performance, we introduce AgriBench-13K Suite, a comprehensive benchmark covering a wide range of agricultural tasks and difficulty levels. AgriGPT also retains strong performance on general-domain benchmarks and demonstrates effective multilingual transfer, reinforcing its robustness beyond agriculture-specific contexts. Together, the model, data engine, dataset, and benchmark form a coherent LLM ecosystem that not only advances agricultural AI research but also enables equitable and practical deployment in real-world, low-resource farming communities—highlighting its broader social impact and its potential to foster inclusive digital transformation in agriculture.

\bibliography{aaai25}

\end{document}